\documentclass[10pt,twocolumn,letterpaper]{article}

\usepackage[pagenumbers]{cvpr} 

\usepackage{graphicx}
\usepackage{amsmath}
\usepackage{amssymb}
\usepackage{booktabs}

\usepackage{cite}
\usepackage{float}
\usepackage{times}
\usepackage{epsfig}
\usepackage{graphicx}
\usepackage{amsmath}
\usepackage{amssymb}
\usepackage{booktabs}
\usepackage[ruled,vlined]{algorithm2e}
\usepackage{url}

\usepackage{array}
\makeatletter
\def\hlinew#1{%
  \noalign{\ifnum0=`}\fi\hrule \@height #1 \futurelet
   \reserved@a\@xhline}
\makeatother

\usepackage{array}
\makeatletter
\newcommand{\thickhline}{%
	\noalign {\ifnum 0=`}\fi \hrule height 1pt
	\futurelet \reserved@a \@xhline
}
\newcolumntype{"}{@{\vrule width 1pt}}
\makeatother

\usepackage{xcolor}

\usepackage[pagebackref,breaklinks,colorlinks]{hyperref}

\usepackage[capitalize]{cleveref}
\crefname{section}{Sec.}{Secs.}
\Crefname{section}{Section}{Sections}
\Crefname{table}{Table}{Tables}
\crefname{table}{Tab.}{Tabs.}

\def\cvprPaperID{5405} 
\def\confName{CVPR}
\def\confYear{2023}

\newcommand{\MCNF}[1]{MCNF}
\newcommand{\ACP}[1]{ACP}
\newcommand{\MNLC}[1]{MNLC}

\begin{document}

\title{Collaborative Noisy Label Cleaner: Learning Scene-aware Trailers for Multi-modal Highlight Detection in Movies}




\author{Bei Gan \quad  Xiujun Shu$^{*}$ \quad Ruizhi Qiao$^{*}$ \quad Haoqian Wu \quad Keyu Chen \quad Hanjun Li \quad Bo Ren\\
Tencent YouTu Lab\\
{\tt\small \{stylegan, xiujunshu, ruizhiqiao, linuswu, yolochen, hanjunli, timren\}@tencent.com}
}

\vspace{-1.2cm}
\twocolumn[{%
\maketitle
\vspace{-20pt}
\begin{figure}[H]
    \hsize=\textwidth
    \centering
    \includegraphics[width=2\linewidth]{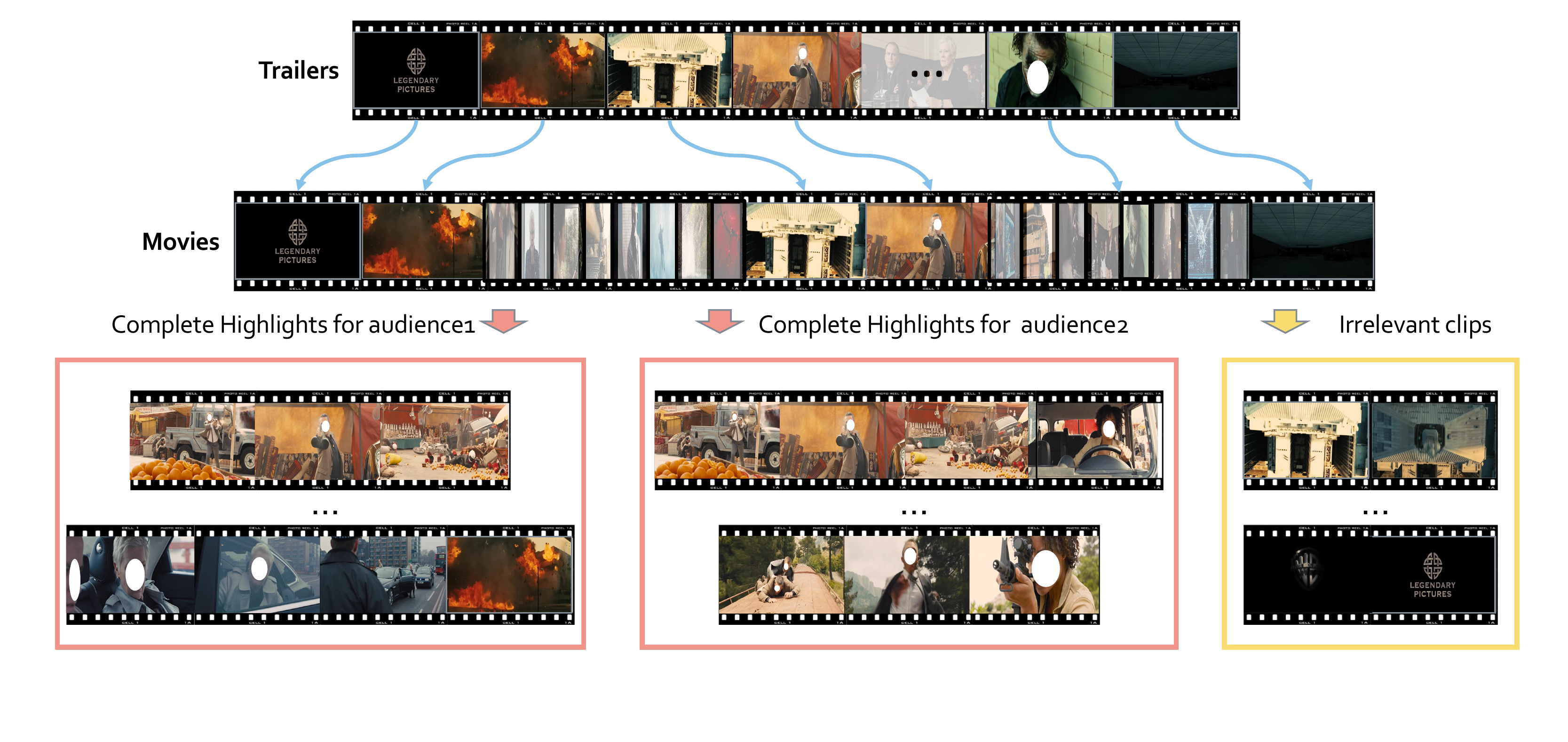}
    \vspace{-1cm}
    \captionof{figure}{
    As the preview of the movie, trailers are selected by professionals to grab an audience’s attention.However, trailers are usually composed with shots sparsely selected from movies to avoid spoilers, and the audience cannot get complete highlight information.Some trailer clips convey the artistic style of the film only and lack movie storylines, disturbing the audience's impressions.In addition, different audiences may be interested in different styles of clips, which makes it challenging to learn highlights from them.}
    \label{fig:trailers}
\end{figure}
}]

{
  \renewcommand{\thefootnote}%
    {\fnsymbol{footnote}}
  \footnotetext[1]{Corresponding author.}
}

\begin{abstract}

    \vspace{-0.2cm}

    Movie highlights stand out of the screenplay for efficient browsing and play a crucial role on social media platforms. Based on existing efforts, this work has two observations: (1) For different annotators, labeling highlight has uncertainty, which leads to inaccurate and time-consuming annotations. (2) Besides previous supervised or unsupervised settings, some existing video corpora can be useful, e.g., trailers, but they are often noisy and incomplete to cover the full highlights. In this work, we study a more practical and promising setting, i.e., reformulating highlight detection as ``learning with noisy labels''. This setting does not require time-consuming manual annotations and can fully utilize existing abundant video corpora. First, based on movie trailers, we leverage scene segmentation to obtain complete shots, which are regarded as noisy labels. Then, we propose a Collaborative noisy Label Cleaner (CLC) framework to learn from noisy highlight moments. CLC consists of two modules: augmented cross-propagation (ACP) and multi-modality cleaning (MMC). The former aims to exploit the closely related audio-visual signals and fuse them to learn unified multi-modal representations. The latter aims to achieve cleaner highlight labels by observing the changes in losses among different modalities. To verify the effectiveness of CLC, we further collect a large-scale highlight dataset named MovieLights. Comprehensive experiments on MovieLights and YouTube Highlights datasets demonstrate the effectiveness of our approach. Code has been made available at:{\url{https://github.com/TencentYoutuResearch/HighlightDetection-CLC}}. 
\end{abstract}
 
\vspace{-0.2cm}

\section{Introduction}
\label{sec:intro}




With the growing number of new publications of movies in theaters and streaming media, audiences become even harder to choose their favorite one to enjoy for the next two hours.
An effective solution is to watch the movie trailers before choosing the right movie. This is because trailers are generally carefully edited by filmmakers and contain the most prominent clips from the original movies.
As a condensed version of full-length movies, trailers are elaborately made with highlight moments to impress the audiences.
Consequently, they are high potential in serving as supervision sources to train automatic video highlight detection algorithms and facilitating the mass production of derivative works for video creators in online video platforms, e.g., YouTube and TikTok.

Existing video highlight detection (VHD) approaches are generally trained with annotated key moments of long-form videos. However, they are not suitable to tackle the movie highlight detection task by directly learning from trailers.
The edited shots in trailers are not equivalent to ground-truth highlight annotations in movies. 
Although a previous work~\cite{LeziWangECCV2020} leverages the officially-released trailers as the weak supervision to train a highlight detector, the highlighted ness of trailer shots is extremely noisy and varies with the preference of audiences, as shown in Fig.~\ref{fig:trailers}. 
On one hand, trailers tend to be purposefully edited to avoid spoilers, thus missing key moments of the storylines.
On the other hand, some less important moments in the original movies are over-emphasized in the trailers because of some artistic or commercial factors.
The subjective nature of trailer shots makes them noisy for the VHD task, which is ignored by existing VHD approaches.

To alleviate the issue, we reformulate the highlight detection task as ``learning with noisy labels''. Specifically, we first leverage a scene-segmentation model to obtain the movie scene boundaries. 
The clips containing trailers and clips from the same scenes as the trailers provide more complete storylines.
They have a higher probability of being highlight moments but still contain some noisy moments. Subsequently, we introduce a framework named Collaborative noisy Label Cleaner (CLC) to learn from these pseudo-noisy labels. 
The framework firstly enhances the modality perceptual consistency via the augmented cross-propagation (ACP) module, which exploits closely related audio-visual signals during training. 
In addition, a multi-modality cleaning (MMC) mechanism is designed to filter out noisy and incomplete labels. 

To support this study and facilitate benchmarking existing methods in this direction, we construct MovieLights, a Movie Highlight Detection Dataset. MovieLights contains 174 movies and the highlight moments are all from officially released trailers. The total length of these videos is over 370 hours. We conduct extensive experiments on MovieLights, in which our CLC exhibits promising results. We also demonstrate that our proposed CLC achieves significant performance-boosting over the state-of-the-art on the public VHD benchmarks. 

In summary, our major contributions are as follows:
\begin{itemize}

  \item We introduce a scene-aware paradigm to learn highlight moments in movies without any manual annotation. To the best of our knowledge, this is the first time that highlights detection is regarded as learning with noisy labels.
  
  \item We present an augmented cross-propagation to capture the interactions across modalities and a consistency loss to maximize the agreement between the different modalities.
  
  \item We incorporate a multi-modality noisy label cleaner to tackle label noise, which further improves the robustness of networks to annotation noise.

  \item Experiments on movie datasets and benchmark datasets validate the effectiveness of our framework.

\end{itemize}

\vspace{-0.2cm}
\section{Related Works}
\label{sec:related_works}

\noindent\textbf{Video Highlight Detection.} 
This task aims to identifying the interesting moments from untrimmed videos. 
In recent years, the videos studied for this task extend from domain-specific sport videos~\cite{Detectinghighlightsinsports} to general videos such as social media videos~\cite{RankingDomainspecificECCV2014}, news~\cite{TVSumCVPR2015}, first-person videos~\cite{HDPDCVPR2016} and vlog~\cite{QVHighlightsNIPS2021} . 
Most of previous works~\cite{Video2gifCVPR2016,HDPDCVPR2016,Threedimensional2018} interpret the video highlight detection task as a segment-level ranking problem. 
They compare pairwise segments from same domain video in order to learn a model that assigns highlight scores to these segments where the highlight segments receive higher scores than the non-highlight segments. 
MINI-Net~\cite{MININetECCV2020} proposes to cast highlight detection as multiple instance ranking network learning. 
SL-Module~\cite{CrosscategoryICCV2021} explores the highlight detection problem through Unsupervised Domain Adaptation (UDA) ~\cite{Asurvey2010}.
UMT~\cite{UMTCVPR2022} integrates highlight detection and moment retrieval into a unified framework and conduct joint optimization.
PLD~\cite{LearningPixelLevelCVPR2022} models the video highlight detection into a  pixel-level distinction estimation task.
In this work, we regard highlight detection as learning with noisy labels.
Joint-VA~\cite{JointVisualandAudioICCV2021} also considers video highlight detection from the perspective of noise.
However, it focuses on noise in features, such as videos having noisy audio when the microphone constantly has water splashing against it. 
We focus on specific annotation noise in video highlight detection. 

\noindent\textbf{Studies on Movies and Trailers.} 
Studies on movies and trailers have received increased attention in research. 
~\cite{MovieNetECCV2020} introduces a comprehensive dataset for movie understanding. 
~\cite{Learninglatent2014,MovieGraphsCVPR2018} try to model the relationships among the movie characters.
~\cite{LGSSCVPR2020,ShotContrastiveCVPR2021,SceneConsistencyCVPR2022} focus on breaking the storylines of movies into semantically cohesive parts.
Besides the studies on movies, efforts have been made to develop trailer understanding.
~\cite{Qualityevaluation2015} presents a movie summarization system and composes movie summaries in terms of user experience evaluation. 
~\cite{MMTF14KACM2018} designs a movie trailer dataset for the evaluation of video-based recommender systems.
~\cite{Fromtrailerstostorylines2018} is the first approach that bridges trailers and movies and allows knowledge learned from trailers to be transferred to full movie analysis.
In ~\cite{LeziWangECCV2020}, the visual module and the temporal analysis module are respectively trained on trailers and movies.
Because of the inaccessibility of public trailer-related benchmarks, we construct a new dataset (MovieLights) to detect the highlight moments in movies.

\noindent\textbf{Learning with Noisy Labels.} 
Learning with noisy labels has been a long-standing problem in computer vision.
There are three kinds of approaches to this problem. 
One of the most common strategies for tackling label noise is to capture the transition probabilities between noisy labels and clean labels~\cite{AreAnchorNIPS2019,ConfidentJAIR2019,PartdependentsNIPS2020,MetaLabelAAAI2021,EstimatingInstanceCVPR2022,PNPCVPR2022}.
Another solution is to design robust loss functions for model training against noisy labels~\cite{GCENIPS2018,SymmetricCrossEntropyICCV2019,DivideMixICLR2020,NormalizedLossICML2020,PeerLossICML2020,LearningwithinstancedependentICLR2021}.
A popular method is to design a mechanism to select clean samples or give lower weight for noise samples in the training set to reduce impact of noise~\cite{ MentorNetICML2018,CoteachingNeurIPS2018,O2UNetICCV2019,AsymmetricCoTeachingAAAI2020,CombatingnoisylabelsCVPR2020,WeblySupervisedICCV2021}. 
In this paper, we attempt to solve the problem by exploiting multi-modalities nature of movies.

\vspace{-0.2cm}

\section{Movie Highlight Dataset}
\label{sec:Movie_Highlight_Dataset}



%
%

In this paper, we aim to detect the highlight moments in movies by learning from easily accessible trailers as the noisy supervision. 
However, the existing movie and trailer-related benchmarks ~\cite{MMTF14KACM2018, CondensedMoviesACCV2020, MovieNetECCV2020} lack sufficient functions for this task, such as the absence of full-length movies and ground-truth highlight annotations.
Huang et al.~\cite{Fromtrailerstostorylines2018} propose to respectively learn visual representations from trailers and temporal structure from full-length movies in their constructed Large-Scale Movie and Trailer Dataset (LSMTD).
Wang~\cite{LeziWangECCV2020}  constructs a Trailer Moment Detection Dataset (TMDD) for detecting trailer moments from full-length movies without explicit human annotation. Both LSMTD and TMDD are not publicly available, while TMDD only contains three movie genres. 

\begin{figure}[t]
   \begin{center}
   \includegraphics[width=1\linewidth]{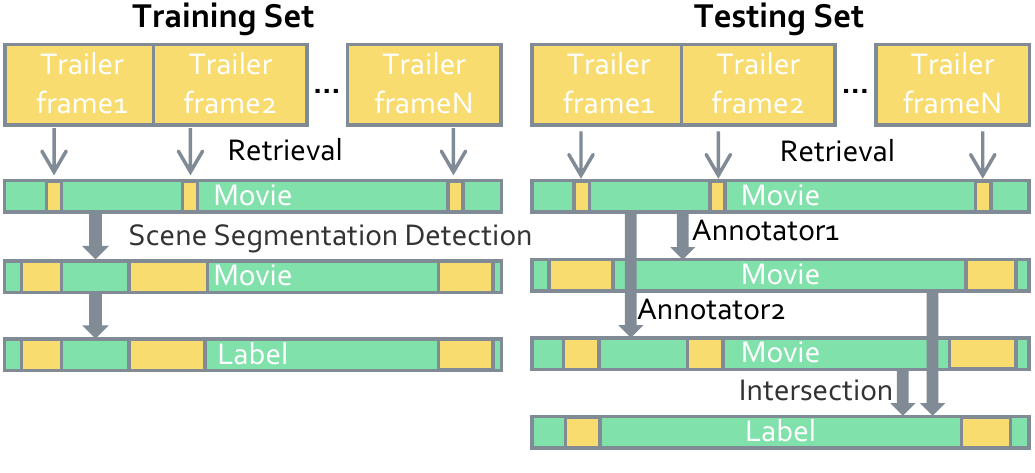}
   \vspace{-0.8cm}
   \end{center}
      \caption{The labeling process of MovieLights. For the training set, we introduce a scene-aware paradigm to obtain labels automatically. For the testing set, we collect $2$ sets of labels from different annotators.}
   \label{fig:annotation}
\end{figure}

%
%
%
%
%

The inaccessibility of public benchmarks motivates us to construct a new dataset, named Movie Highlight Detection Dataset (MovieLights). 
In particular, we purchase a set of movies from commercial channels and collect their corresponding trailers from streaming platforms such as YouTube, covering at least $25$ genres to ensure content diversity.
The movies and trailers are then prepossessed by shot segmentation ~\cite{TransNetV2} and scene segmentation ~\cite{SceneConsistencyCVPR2022}, respectively. 
The resulting shots are a series of consecutive frames taken by the camera until a physical interruption, and the scenes are consecutive shots that share a semantically related theme. 

As seen in Fig.~\ref{fig:annotation}, to build the ground truth, we conduct Faiss~\cite{Faiss2019} to obtain visual similarity matching between trailer frames and movie frames.  We locate trailer moments in the movie and align them with the movie shots as annotation references.
For the testing set, we collect $2$ sets of moments for each movie from different workers, and these moments are annotated by different annotators independently.
To ensure the consistency of results from different annotations, during the annotation procedure, all highlight moments must be related to the annotation references.
Though all selected shots are relevant to the trailers, as highlight moments can be subjective, they may still vary in their saliency and time span.
We calculate the intersection between every pair of moments annotated as the ground truth.
However, the vast diversity of movie storylines makes the annotation challenging as it is time-consuming and requires annotators to be familiar with the movie. 
To collect a large amount of training data efficiently, we introduce a scene-aware paradigm to obtain the highlight moments label without any manual annotation.
Specifically, we expand the shot-level annotation references to the scene span as positive samples automatically.
It will capture the complete scene context of the trailer shot with movie storylines.
Since the trailer shots may contain some less important moments, the acquired highlight labels are still noisy. 

\begin{table}[]
   \caption{
   The basic statistics of MovieLights.
   }

   \vspace{-0.3cm}
   \label{tab:dataset}
   \small
   \centering
   \begin{tabular}{c|cc}
   \hlinew{1.1pt}
     & \multicolumn{1}{l}{Train} & \multicolumn{1}{l}{Test}  \\\hline\hline
    Movie Number                            & 144                       & 30                        \\
    Avg Durations per Movie                       & \multicolumn{1}{l}{2.19h} & \multicolumn{1}{l}{2.14h} \\
    Avg Shot Number per Movie                            & 1852               & 1940           \\
    Avg Scene Number per Movie                           & 207              & 193              \\
    Annotator1 Positive sample Proportion & -          & 0.27                      \\
    Annotator2 Positive sample Proportion & -          & 0.30                      \\
    Positive sample proportion            & 0.35                      & 0.21     \\     
    \hlinew{1.1pt}
   \end{tabular}
\end{table}

As seen in Tab.~\ref{tab:dataset}, MovieLights contains $174$ movies in full length with their official trailers and it is split into a training set with $144$ movies, and a testing set with $30$ movies. 
%
The content diversity is ensured by the rich domain informations (more than $25$ genres) and abundant segments ($325k$  shots and $36k$ scenes). 
Most movies in our dataset are between $90$ to $150$ minutes, and the length of the annotated moments varies from tens of seconds to several minutes. 
As the acquired positive highlight moments take up $35\%$ in the training set while the annotated true positives take up $21\%$ in the testing set, the difference tells the obvious existence of label noise. 

We plan to release the dataset publicly to promote further study of movie analysis. Due to copyright issues, trailers and movies will be released in the form of extracted features in visual and audio modalities. 

\vspace{-0.2cm}

\section{Approach}
\label{sec:approch}

\subsection{Overview}
\label{subsec:Overview.}

\begin{figure*}[t]
   \begin{center}
   \includegraphics[width=1.0\linewidth]{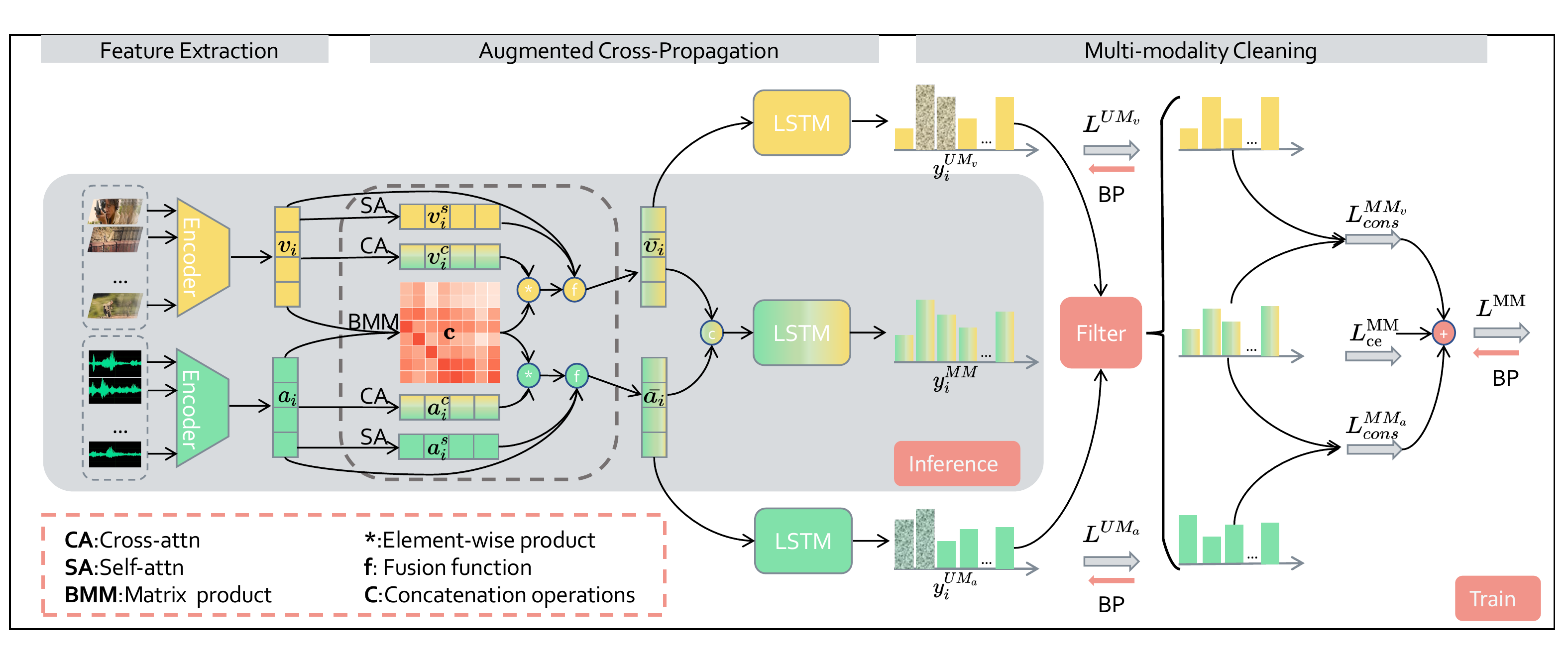}
   \end{center}
      \vspace{-0.5cm}
      \caption{
         Overview of the proposed CLC. It includes three modules: feature extraction, augmented cross-propagation (ACP), and multi-modal cleaning (MMC). The visual and audio modalities of the input video are represented as vectors by the feature extraction module. Then the features are augmented by ACP module to capture semantic associations acorss modalities. MMC is used to filter outs noisy and incomplete labeling with additional uni-modal branches. During inference, we remove the uni-modal branches and only rely on the prediction of multi-modal branch. More details of the CLC are shown in Sec.~\ref{subsec:Overview.}}
      \vspace{-0.5cm}
   \label{fig:pipeline}
\end{figure*}

The overall architecture of our Collaborative noisy Label Cleaner (CLC) framework is illustrated in Fig.~\ref{fig:pipeline}, which includes three modules: feature extraction, augmented cross-propagation (ACP), and multi-modal cleaning (MMC). 

In our framework, both the visual and audio modalities are utilized. 
For feature extraction and encoding, we first split the video $V$ into $T$ shots. We characterize the $i^{th}$ shots by two vectors, \emph{i.e.,} $\mathbf{v}_i$ for the visual features, and $\mathbf{\hat{a}_i}$ for the audio features, where $i = 1, 2, ..., T$. These features are extracted using pre-trained visual~\cite{dosovitskiy2020image} and audio feature extractors~\cite{kong2020panns}. The parameters of the two extractors are frozen during training. 

The ACP and MMC are the core components of our framework. Since the visual-audio signals in videos are closely related but do not always contribute to highlight detection, the ACP module exploits the relationship via uni-modal and cross-modal interactions. Then, it learns unified multi-modal representations for highlight detection. As the obtained highlight moments after scene segmentation are noisy, the MMC module firstly observes the changes in uni-modal losses, then filters the noisy labels and utilizes the clean ones for multi-modal supervised learning. Next, we will introduce details for each component.






\subsection{Augmented Cross-Propagation}
\label{subsec:Augmented Cross-Propagation.}
\noindent\textbf{Cross-Propagation.} 
To predict the highlights, the model needs to understand the storylines of the movie. Meanwhile, visual and audio inputs do not always contribute to accurate prediction. Therefore, temporal modeling and modality interaction are the keys to achieving successful highlight detection. To achieve this goal, we design our ACP module to fuse the visual-audio modalities. It involves three steps in total.



First, in order to align the multi-modal features, we introduce $h$ which is fully-connected (FC) layer with ReLU to the $\mathbf{\hat{a}_i}$ such that it has same dimension as the $\mathbf{v}_i$.
\begin{equation}
\mathbf{a}_i=h(\mathbf{\hat{a}_i}).
\end{equation}

Movies are composed of consecutive audio-visual clips. 
Therefore, to measure the highlighted ness of a given shot, one must consider the relationship of the shot with its adjacent shots. 
The self-attention mechanism has shown effectiveness in capturing the long-term dependencies in previous works. 
We leverage self-attention to capture the temporal relationship for $\mathbf{v}_i$ and $\mathbf{a}_i$ via Eq.~\ref{eq:self-attention-v} and Eq.~\ref{eq:self-attention-a}, respectively. 
\begin{equation}
\mathbf{v}_{i}^{s}=softmax\left(\frac{(\mathbf{v}_{i}{W}_{1}^{v})( \mathbf{v}{W}_{2}^{v})^{\top}}{\sqrt{d}}\right) (\mathbf{v}{W}_{3}^{v}),
\label{eq:self-attention-v}
\end{equation}
\begin{equation}
\mathbf{a}_{i}^{s}=softmax\left(\frac{(\mathbf{a}_{i}{W}_{1}^{a})( \mathbf{a}{W}_{2}^{a})^{\top}}{\sqrt{d}}\right) (\mathbf{a}{W}_{3}^{a}),
\label{eq:self-attention-a}
\end{equation}
where $\mathbf{v}=[\mathbf{v}_{1};\mathbf{v}_{2};...;\mathbf{v}_{T}]$ and $\mathbf{a}=[\mathbf{a}_{1};\mathbf{a}_{2};...;\mathbf{a}_{T}]$;
the scaling factor $d$ is equal to the visual/audio feature dimension and $(*)^{\top}$ denotes the transpose operator;
$W^v$ and $W^a$ are learnable matrices of two modalities, which are implemented by a linear layer.
Uni-modal self-attention can well capture uni-modal temporal contexts and enhance clip features within the same modality.

Despite the above self-attention capturing the clip interactions within the uni-modality, it is critical to capture the interactions across modalities. 
To capture semantic associations based on multi-modal signals, we introduce cross-attention to update the features of each modality.
\begin{equation}
\mathbf{v}_{i}^{c}=softmax\left(\frac{(\mathbf{v}_{i}{W}_{4}^{v}) (\mathbf{a}{W}_{4}^{a})^{\top}}{\sqrt{d}}\right) (\mathbf{a}{W}_{5}^{a}),
\label{eq:cross-attention-v}
\end{equation}
\begin{equation}
\mathbf{a}_{i}^{c}=softmax\left(\frac{(\mathbf{a}_{i}{W}_{6}^{a}) (\mathbf{v}{W}_{5}^{v})^{\top}}{\sqrt{d}}\right) (\mathbf{v}{W}_{6}^{v}),
\label{eq:cross-attention-a}
\end{equation}

Through cross-attention, the information from two modalities are connected. 
However, considering audio-visual temporal asynchrony, it is necessary to select effective information from multi-modality. 
We augment the relevant positive connections and dampen the irrelevant connections.
The strength of these connections is measured by the cross-correlation matrix, computed by,
\begin{equation}
\mathbf{c}^v=ReLU\left(\frac{\mathbf{v}\mathbf{a}^{\top}}{\sqrt{d}}\right),
\mathbf{c}^a=ReLU\left(\frac{\mathbf{a}\mathbf{v}^{\top}}{\sqrt{d}}\right).
\label{eq:cross-correlation}
\end{equation}


For each modality, the cross-correlation matrix $\mathbf{c}$ is used to re-weight the cross-attention features.
Finally, we obtain updated visual features $\mathbf{\bar{{v}_i}}$ and audio features $\mathbf{\bar{{a}_i}}$ by fusion of the original features, enhanced uni-modal features, and cross-modal features.
\begin{equation}
\mathbf{\bar{{v}_i}}=f(\mathbf{v}_i, \mathbf{v}_i^s, (\sum_{j=1}^{T} \mathbf{c}_{ij}^{v}) * \mathbf{v}_i^c),
\end{equation}
\begin{equation}
\mathbf{\bar{{a}_i}}=f(\mathbf{a}_i, \mathbf{a}_i^s, (\sum_{j=1}^{T} \mathbf{c}_{ij}^{a}) * \mathbf{a}_i^c),
\end{equation}
%
where $f$ is the fusion function consisting of FC layers and ReLU to further project the features. 

Clearly, the cross-correlation matrix will assign large weights to clips that are relevant to the other modality. 
In addition, the ReLU activation in 
Eq.~\ref{eq:cross-correlation}
cuts off connections with negative similarity values and only relevant positive connections would be preserved. 
By the above operations, the original feature will be infused with richer information. 



\noindent\textbf{Consistency Loss.} 
Multi-modal inputs help to comprehensively learn by integrating different aspects and boosting model performance. 
However, they are not fully exploited because some modality-specific features may be weakened in the fusion even when the multi-modal model outperforms its uni-modal counterpart. 
In this work, we provide parallel branches for each modality separately to obtain uni-modal prediction during training. 
As seen in Fig.~\ref{fig:pipeline}, all branches share the same clip feature extracted from ACP and the feature is fed into different branches independently. In each branch, we develop a temporal model $G$ to obtain its prediction score $\mathbf{y}_i$ of the $i^{th}$ clip being a highlight as follows:
\begin{equation}
\mathbf{y}_i^{MM}=G(\mathbf{\bar{{v}_i}},\mathbf{\bar{{a}_i}}),\mathbf{y}_i^{UM_v}=G(\mathbf{\bar{{v}_i}}),\mathbf{y}_i^{UM_a}=G(\mathbf{\bar{{a}_i}}).
\end{equation}
where ${MM}$ is multi-modal branch; ${UM_v}$ is visual branch and ${UM_a}$ is audio branch.

However, the gap between the different modalities would bring instability in the joint optimization process. 
We employ the auxiliary consistency loss to guarantee consistency between the different modalities.
Given the multi-modal features of a clip, they are consistent if they share the same prediction. 
Specifically, the consistency loss is defined as the cross-entropy between the multi-modal prediction probability $\mathbf{y}_i^{MM}$ and uni-modal prediction probability $\mathbf{y}_i^{UM}$ :
\begin{equation}
\footnotesize
\mathcal{L}_{\text {cons }}=-\left(\sum_{i=1}^N \mathbf{y}_i^{UM_v} \log \mathbf{y}_i^{MM} + \sum_{i=1}^N \mathbf{y}_i^{UM_a} \log \mathbf{y}_i^{MM}\right),
\end{equation}
%
where $N$ is the number of samples in a batch.

This consistency loss not only implicitly enhances uni-modal information, but also explicitly guides the multi-modal branch to robuster supervision. 
The added uni-modal branches are only utilized in the training phase and are disabled during the inference stage. 





\subsection{Multi-modal Cleaning}
\label{subsec:Multi-modality Noisy Label Cleaner.}

\noindent\textbf{Multi-modality Sample Cleaning.} 
Annotation noise is inevitable in VHD due to subjectivity depending on the users and annotators. 
In this paper, we argue that highlight detection should be regarded as Learning with Noisy Labels.
To alleviate the performance drop caused by noisy labels, we adopt a multi-modality collaborative cleaning to adaptively filter noisy samples with noisy modality information.


Firstly, we maintain multiple outputs simultaneously which are predicted by different branches.
The uni-modal branches independently select clean samples based on the low-loss criterion in which instances with lower losses are treated clean samples. 
Contrary to existing noisy sample selection methods~\cite{ExtractingUsefulMM2021, ConfidentLearningJAIR2021}, which directly discard high-loss samples, we keep all samples to train the uni-modal branches.
Then, we update the multi-modal branch using only clean samples selected by both uni-modal branches in the back-propagation.
The samples for the multi-modal branch training are selected dynamically while all samples participate in the uni-mode branches training.
In this way, MMC obtains information on all samples to avoid the model defecting to favoring easy samples.
More details of MMC are presented in the supplementary material. 


Each branch has its own loss function. 
For the uni-modal branch, we employ the cross-entropy loss with all samples as follows: 
\begin{equation}
\mathcal{L}^{UM_v}=\mathcal{L}_{\text {ce }}^{UM_v}=-\sum_{i=1}^N \mathbf{g}_i \log \mathbf{y}_i^{UM_v},
\end{equation}
\begin{equation}
\mathcal{L}^{UM_a}=\mathcal{L}_{\text {ce }}^{UM_a}=-\sum_{i=1}^N \mathbf{g}_i \log \mathbf{y}_i^{UM_a}.
\end{equation}

For the multi-modal branch, its cross-entropy loss and consistency loss are updated with a re-weighting scheme with clean samples. 
\begin{equation}
\mathcal{L}_{\text {ce }}^{\text {MM }}=-\sum_{i=1}^{N^{\prime}} \mathbf{g}_i \log \mathbf{y}_i^{MM},
\end{equation}
\begin{equation}
\footnotesize
\mathcal{L}_{\text {cons }}^{MM }=-\left(\sum_{i=1}^{N^{\prime}} \mathbf{y}_i^{UM_v} \log \mathbf{y}_i^{MM} + \sum_{i=1}^{N^{\prime}} \mathbf{y}_i^{UM_a} \log \mathbf{y}_i^{MM}\right),
\end{equation}
\begin{equation}
\mathcal{L}^{\text {MM }}=\mathcal{L}_{\text {ce }}^{\text {MM }}+\beta \mathcal{L}_{\text {cons }}^{MM },
\label{eq:LOSS}
\end{equation}
where $\mathbf{g}_i$ and $\mathbf{y}_i^{*}$ denote the ground-truth and predicted probability of the $i^{th}$ clip, respectively; $N$ is the number of samples in a batch and ${N^{\prime}}$ is the number of samples seleted by MMC;
and $\beta$ are designed to balance different loss terms.

\noindent\textbf{Post processing.} 
In experiments we observe that noise makes jitter prediction curves along the temporal dimension, which may cause discontinuous thresholds for highlight selection. 
Therefore, we apply a median filter to smooth the prediction curves. Supposing 
$\mathbf{y}=[\mathbf{y}_{1};\mathbf{y}_{2};...;\mathbf{y}_{T}]$ is the original curve predicted by the CLC, the smoothed curve 
$\mathbf{s}=[\mathbf{s}_{1};\mathbf{s}_{2};...;\mathbf{s}_{T}]$  is given by:
\begin{equation}
\mathbf{s}_i= \begin{cases}\operatorname{Med}\left(\mathbf{y}_{i-k}, \mathbf{y}_{i+k}\right), & k<i \leq T-k \\ \mathbf{y}_i, & \text { otherwise }\end{cases}
\label{eq:median-filter}
\end{equation}
where $k$ is the window size, and ``Med'' denotes the median filter.

\vspace{-0.2cm}

\section{Experiment}
\label{sec:experiment}

\subsection{Datasets and Experimental Settings}
\label{subsec:Datasets and Experimental Settings}

\noindent\textbf{Datasets.}
\label{subsubsection:Datasets.}
We evaluate our CLC on the constructed dataset MovieLights and public YouTube Highlights dataset~\cite{RankingDomainspecificECCV2014}. 
MovieLights is split into training and testing sets, each containing 144 and 30 movies respectively. 

We split movies into shots. The movie features are represented at shot-level. 
We use the middle frame of each shot to extract its visual feature with ViT~\cite{VITICLR2021} pre-trained by CLIP~\cite{CLIP2021}. 
We align timestamps of audio clips with the visual shots, and sample the audio clip of each with $16K$ Hz sampling rate and $512$ windowed signal length. The resulted shot-level audio features are obtained with the PANN audio network~\cite{kong2020panns} pretrained on AudioSet~\cite{Audioset2017}. 

The YouTube Highlights contains six distinct categories with a total 422 videos currently available. 
Following the practice of prior efforts, we train a highlight detector for each category. 
%
YouTube Highlights provides two annotations: Harvested Highlight and Mturk Highlight. 
In the Harvested annotation, the match label specifies if each clip is matched in the edited video, where 1 denotes matched, -1 denotes unmatched and 0 denotes the borderline cases.
In the Mturk annotation, the highlight labels are marked by multiple turkers of different styles, making it noisier than the Harvested annotation. 


On YouTube Highlights, we use the same protocol and data preprocessing as ~\cite{UMTCVPR2022}.
It obtain clip-level visual features and optical flow features using I3D~\cite{I3DCVPR2017} pre-trained on Kinetics400~\cite{K400}.
It use a PANN audio network~\cite{kong2020panns} pretrained on AudioSet~\cite{Audioset2017} to obtain audio features that align with the visual clips. 
Frame-level features are average-pooled within each clip for both audio and visual features to generate a clip-level feature. 
Since each feature vector spans 32 consecutive frames, we follow ~\cite{UMTCVPR2022} and consider the feature vector corresponded to a clip if their overlap is more than $50\%$.  


\noindent\textbf{Benchmarks.}
\label{subsubsection: Benchmarks.}
To better inspect the robustness of our CLC against noisy labels, we also apply label perturbations in training set of YouTube Highlights while keeping the ground-truths in the testing set unchanged.
YouTube Highlights has two benchmarks for comparison. 
1) \textbf{Harvested with matched:} 
we regard the clips labeled with matched as the highlighted clips and this is the same setting as in previous works ~\cite{UMTCVPR2022,CrosscategoryICCV2021}. 
2) \textbf{Harvested with matched and borderline:}
clips labeled with matched and borderline are treated as the highlight moments.
In this benchmark, the training set contains some highlighted clips with weak confidence. 
We select the clips whose mturk-label is over score 1 as the highlighted clips, which means that at least one turker selects the clip as a highlight. 
There are labeled by different types of annotators between Mturk and the test set and bring greater noise. 
We select the clips whose mturk-label is over score 1 as the highlighted clips, which means that at least one turker selects the clip as a highlight. Due the labeling gap between the Mturk annotation in the training set and the clean testing set, this benchmark is even noisier.

\noindent\textbf{Baselines.}
\label{subsubsection:Baselines.}
We introduce CLC$-$, the degenerated version of CLC, as a baseline.
Similar to CLC, CLC$-$ is a Bi-LSTM-based model and takes the temporal sequence of shot features as input but lacks the modules of augmented cross-propagation and multi-modality noisy label cleaner. 

\noindent\textbf{Evaluation Metric.}
\label{subsubsection: Evaluation Metric.}
We adopt mean Average Precision (mAP) as the evaluation metric for MovieLights and YouTube Highlights. 
Considering that a highlighted moment in one video is not necessarily more interesting than non-highlight moments in other videos, we evaluate on each test video independently and report the averaged results.


\noindent\textbf{Implementation Details.}
\label{subsec: Implementation Details.}
On the MovieLights, we train our model using SGD, with a learning rate of 0.01. We train for 50 epochs. Before the cross-attention modules, we project each modality into a vector of 512 dimension. The key, query, and value vectors all share the same dimension.
Weight $\beta$ in Eq.~\ref{eq:LOSS} is empirically set to 0.1 and window size $k$ in Eq.~\ref{eq:median-filter} is set to 9.
 
\subsection{Results on MovieLights}
\label{subsec:Results on MovieLights.}

\begin{table}[!t]
    \caption{Results on MovieLights.}
    \small
	\centering  
	\label{tab:moviesota}
	\renewcommand\arraystretch{1} 
	\vspace{-0.3cm}
    \begin{tabular}{p{2.5cm}p{1.5cm}<{\centering}|p{1.5cm}<{\centering}} 
   \hlinew{1.1pt}
    Methods    & Modality & mAP   \\ 
    \hline \hline
    GIFs~\cite{Video2gifCVPR2016}       & V       & 25.48\\ 
    SL-Module~\cite{CrosscategoryICCV2021}       & V       & 32.34 \\ 
    SL-Module~\cite{CrosscategoryICCV2021}      & VA       & 34.27 \\ 
    UMT~\cite{UMTCVPR2022}        & VA       & 38.7 \\\hline
    CLC$-$       & VA       & 39.65 \\ 
    CLC$-$ w/ SCE~\cite{SymmetricCrossEntropyICCV2019} & VA       & 39.83                     \\ 
    CLC$-$ w/ LS~\cite{LabelsmoothingCVPR2016} & VA       & 40.49 \\\hline
    CLC       & VA       & \textbf{43.88}                     \\ 
    \hlinew{1.1pt}
   \end{tabular}
\end{table}


On MovieLights, we train our model with the noisy pseudo labels. 
To compare with previous state-of-the-art highlight detection works, we train UMT~\cite{UMTCVPR2022} and  SL-Module~\cite{CrosscategoryICCV2021} using the same protocol and data preprocessing as in CLC. We also compare with Video2GIF~\cite{Video2gifCVPR2016} using its off-the-shelf tool\footnote{\href{https://github.com/gyglim/video2gif_code}{https://github.com/gyglim/video2gif\_code}}. The upper part of Tab.~\ref{tab:moviesota} illustrates the significant performance gain of CLC. 


To demonstrate the advantages of CLC in learning with noisy labels,  we make comparisons with two main-stream label noise approaches: Label Smoothing~\cite{LabelsmoothingCVPR2016} and SCE loss~\cite{SymmetricCrossEntropyICCV2019}.
Specifically, we insert Label Smoothing or SCE into our CLC$-$ framework to create two baseline VHD methods to tackle label noise.
The bottom part of Tab.~\ref{tab:moviesota} shows that CLC outperforms the two baseline methods by a notable margin, indicating that our augmented cross-propagation and multi-modality noisy label cleaner are more effective than vanilla label noise approaches in VHD tasks. 

\subsection{Ablation Study}
\label{subsec:Ablation Study.}






\begin{table}[!t]
    \caption{Ablation results of MoiveLighgts.}
    \small
	\centering  
	\label{tab:ablationstudy}
	\renewcommand\arraystretch{1} 
	\vspace{-0.3cm}
    \begin{tabular}{p{0.8cm}<{\centering}p{0.8cm}<{\centering}p{0.8cm}<{\centering}p{0.8cm}<{\centering}|p{1.5cm}<{\centering}} 
   \hlinew{1.1pt} 
    MMSC & CP & CL & PP  & mAP \\\hline \hline
    ×                      & ×               & ×                & ×                & 39.65                   \\
    \checkmark\                      & ×               & ×                & ×                & 41.69                   \\
    \checkmark                      & \checkmark               & ×                & ×                & 42.79                   \\
    \checkmark                      & \checkmark               & \checkmark                & ×                & 43.22                   \\
    \checkmark                      & \checkmark               & \checkmark                & \checkmark                & \textbf{43.88}       \\
    \hlinew{1.1pt}
  \end{tabular}
\end{table}

In this experiment, we analyze the impact of each module. The results are summaried in Tab.~\ref{tab:ablationstudy}.

\noindent\textbf{Multi-modality Sample Cleaning.}
We first inspect the impact of the multi-modality sample filter module because we are primary concerned with learning with noisy labels in VHD. 
The $2\%$ performance gain from the module over baseline shows the importance of filtering noisy sample. 
Based on the this observation, the subsequent ablation experiments are conducted under the setting of noise filtering. 

\noindent\textbf{Cross-Propagation.}
We then evaluate the impact of the feature augmentation module.
Compared to naive feature concatenation, the models with augmented features show superior performance.
The results validate that feature augmentation module can better explore the complementary information from different modalities and suppress the mutual disturbance of desynchronized uni-modal information. 

\noindent\textbf{Consistency Loss.}
We examine the contribution of the consistency loss.
The results demonstrate that the employment of the auxiliary multi-modal constraint further increases the model robustness.
This consistency loss not only implicitly enhances uni-modal information, but also explicitly guides the multi-modal branch to better learning.

\begin{figure*}[t]
   \begin{center}
   \includegraphics[width=1.0\linewidth]{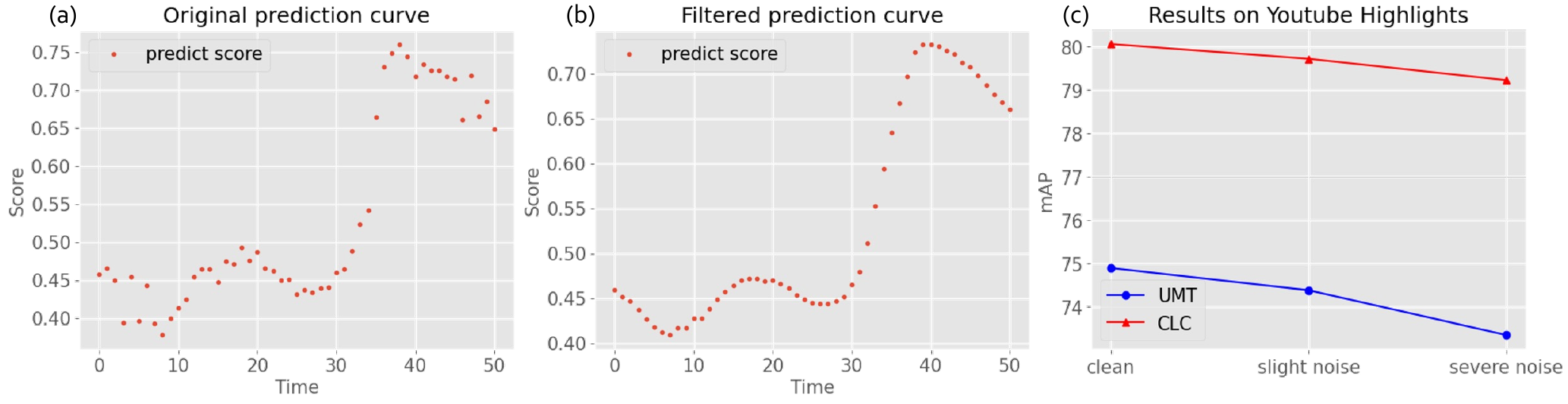}
   \end{center}
      \vspace{-0.4cm}
      \caption{
         (a, b) Comparison of original prediction curve with filtered prediction curve.
         (c) Results on YouTube Highlights with noisy label.}
      \vspace{-0.3cm}
   \label{fig:filter_and_noise}
\end{figure*}

\noindent\textbf{Post Processing.}
We compare the highlight prediction curves with and without the post processing median filter in Fig.~\ref{fig:filter_and_noise}(a,b), which prevents disruptive prediction variation and improves the overall performance. 

\subsection{Results on YouTube Highlights}
\label{subsec:Results on Youtube Highlights.}

\begin{table}[!t]
    \caption{
     Results on YouTube Highlights.
   } 
    \small
	\centering  
	\label{tab:youtubehighligh}
	\renewcommand\arraystretch{1} 
	\vspace{-0.3cm}
    \begin{tabular}{p{2.1cm}|p{0.42cm}<{\centering}p{0.42cm}<{\centering}p{0.42cm}<{\centering}p{0.42cm}<{\centering}p{0.42cm}<{\centering}p{0.42cm}<{\centering}p{0.44cm}<{\centering}} 
    \hlinew{1.1pt} 
    Methods      & dog & gym. & park. & ska. & ski. & surf. & Avg. \\ \hline \hline
    GIFs~\cite{Video2gifCVPR2016}            & 30.8                    & 33.5                     & 54                        & 55.4                     & 32.8                     & 54.1                      & 46.4                     \\
    LSVM~\cite{RankingDomainspecificECCV2014}           & 60.0                    & 41.0                     & 61.0                      & 62.0                     & 36.0                     & 61.0                      & 53.6                     \\
    HighlightMe~\cite{HighlightMeICCV2021}           & 63                      & 73                       & 72                        & 64                       & 52                       & 62                        & 64                       \\
    MINI-Net~\cite{MININetECCV2020}           & 58.2                    & 61.7                     & 70.2                      & 72.2                     & 58.7                     & 60.1                      & 64.4                     \\
    CHD~\cite{ContrastiveLearningCVPR2022}           & 60.6                    & 71.1                     & 74.2                      & 49.8                     & 68.2                     & 68.5                      & 65.4                     \\
    Trail~\cite{LeziWangECCV2020}             & 63.3                    & 82.5                     & 62.3                      & 52.9                     & 74.5                     & 79.3                      & 69.1                     \\
    SL-Module~\cite{CrosscategoryICCV2021}           & 70.8                    & 53.2                     & 77.2                      & 72.5                     & 66.1                     & 76.2                      & 69.3                     \\
    Joint-VA~\cite{JointVisualandAudioICCV2021}          & 64.5                    & 71.9                     & 80.8                      & 62                       & 73.2                     & 78.3                      & 71.8                     \\
    PLD~\cite{LearningPixelLevelCVPR2022}          & 74.9                    & 70.2                     & 77.9                      & 57.5                     & 70.7                     & 79                        & 73                       \\
    CO-AV~\cite{ProbingVisualAudio2022}          & 60.9                    & 66                       & 89                        & 74.1                     & 69                       & 81.1                      & 74.7                     \\
    UMT~\cite{UMTCVPR2022}            & 65.9                    & 75.2                     & 81.6                      & 71.8                     & 72.3                     & 82.7                      & 74.9                     \\
    \textbf{CLC}(ours) & 70.5                    & 79.4                     & 83.9                      & 83.5                     & 79.5                     & 83.6                      & \textbf{80.1}      \\
    \hlinew{1.1pt}
   \end{tabular}
\end{table}

\begin{table*}[]
   \caption{
   Results on YouTube Highlights with Noisy Label.
   }

  \vspace{-0.3cm}
   \label{tab:noisyyoutubehighligh}
   \small
   \centering
    \begin{tabular}{ll|l|rrrrrrl}
    \hlinew{1.1pt}
    Annotation & Noise        & Methods & \multicolumn{1}{l}{dog} & \multicolumn{1}{l}{gym.} & \multicolumn{1}{l}{park.} & \multicolumn{1}{l}{ska.} & \multicolumn{1}{l}{ski.} & \multicolumn{1}{l}{surf.} & \multicolumn{1}{l}{Avg.}         \\\hline\hline
    Harvested matched    & clean        & UMT~\cite{UMTCVPR2022}      & 65.90                   & 75.20                    & 81.60                     & 71.80                    & 72.30                    & 82.70                     & 74.90                                            \\
    Harvested matched    & clean        & CLC    & 70.51                   & 79.43                    & 83.85                     & 83.51                    & 79.46                    & 83.56                     & 80.05 ($\uparrow5.15$)            \\\hline
    Harvested borderline & slight noise & UMT~\cite{UMTCVPR2022}      & 65.93                   & 74.31                    & 81.58                     & 71.84                    & 70.24                    & 82.46                     & 74.39                                           \\
    Harvested borderline & slight noise & CLC    & 69.41                   & 80.73                    & 78.50                     & 85.36                    & 81.11                    & 83.16                     & 79.71 ($\uparrow5.32$)                \\\hline
    Mturk      & severe noise & UMT~\cite{UMTCVPR2022}      & 63.78                   & 76.16           & 75.02                     & 73.62                    & 69.99                    & 81.59                     & 73.36                                              \\
    Mturk      & severe noise & CLC    & 66.92                   & 80.44                    & 85.92                     & 82.33                    & 78.05                    & 81.72                     & 79.22 ($\uparrow5.86$)                     \\ 
    \hlinew{1.1pt}
\end{tabular}
\end{table*}


\begin{figure*}[t!]
   \begin{center}
   \includegraphics[width=1.0\linewidth]{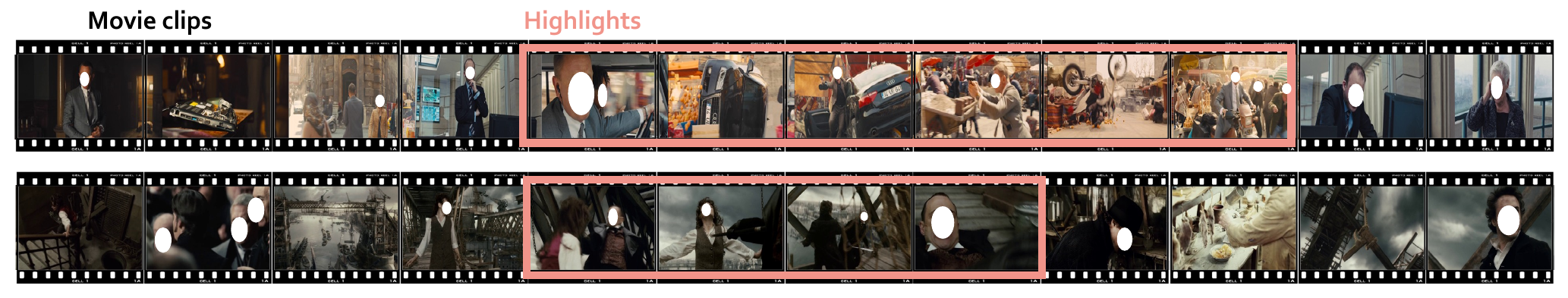}
   \end{center}
      \vspace{-0.5cm}
      \caption{
         The highlight moments selected by our CLC from Skyfall and Sherlock Holmes. Top: Bond gives chase to a professional hitman by car to find a classified hard drive, and then a firefight erupted in the market. Bottom: Holmes and Blackwood are facing off, and then Blackwood reaches for a weapon to kill Holmes, but accidentally trips off a scaffolding and falls to his death.}
      \vspace{-0.5cm}
   \label{fig:visualization}
\end{figure*}



We conduct experiments on the public video highlight detection benchmarks YouTube Highlights, including six domain video datasets, to verify the generalization ability of CLC. 


The setting of \textbf{Harvested Highlight with matched} is consistent with the previous work~\cite{UMTCVPR2022}.
As shown in Tab.~\ref{tab:youtubehighligh}, CLC achieves state-of-the-art performance on the YouTube Highlights, outperforming the existing multi-modal highlight detection methods in the average metric across all categories.
Specifically, CLC achieves best performance in three out of the six categories, while maintaining reasonably competitive performance in the other three categories. 
These results support our claim that  highlight detection should be regarded as learning with noisy labels.



To quantify how CLC is robust to different levels of label noise, we inspect the settings of \textbf{Harvested Highlight with matched and borderline} and \textbf{Mturk Highlight}, which are perturbed by varying degrees of label noise in YouTube Highlights. 


Tab.~\ref{tab:noisyyoutubehighligh} exhibits the performances of CLC and UMT~\cite{UMTCVPR2022} at different noise levels.
As the noise level increases, the VHD task becomes more difficult, but the performance superiority of our CLC over UMT becomes even more obvious. 
It is illustrated in Fig.~\ref{fig:filter_and_noise}(c) that compared with the most recent state-of-the-art UMT, our model can achieve better performance even when disturbed by severe label noise. 




\subsection{Visualization}
\label{Visualization}


As shown in Fig.~\ref{fig:visualization}, we present some visualization examples of the detected highlight clips in MovieLights by CLC. The examples clearly shows that the prediction of CLC is in accordance with user expectation. 
We will provide more examples in the supplementary material.

\vspace{-0.2cm}
\section{Conclusion}
\label{sec:conclusion}


In this study, we present Collaborative noisy Label Cleaner (CLC), a novel framework to handle noisy labels in video highlight detection.
We make use of the augmented cross-propagation module to better enhance network robustness and multi-modality cleaning to achieve cleaner highlight labels by observing the loss changes of different modalities.
We demonstrate the state-of-the-art performance of our method with extensive experiments on MovieLights and YouTube Highlights datasets.
In future work, we are interested in extending the proposed mechanisms to other video-understanding tasks such as scene segmentation and video temporal grounding.

\vspace{-0.3cm}

\begin{algorithm*}[h]
    \DontPrintSemicolon
    \SetAlgoLined
    \SetKwInput{KwInput}{Input}
    \SetKwInput{KwOutput}{Output}
    \KwInput{
    Training dataset $D$,
    a multi-modal branch $G^{MM}$, two uni-modal branch $G^{{UM}_v}$ and $G^{{UM}_a}$,  clean sample proportion $\tau\in[0,1]$, iteration $I_{max}$, epoch $E_{max}$}
    \KwOutput{Updated branch $G^{MM}$, $G^{{UM}_v}$ and $G^{{UM}_a}$ }
    \For{$E=1,2,...,E_{max}$}{
        \For{$I=1,2,...,I_{max}$}{
            Sample a mini-batch $N$ from the dataset $D$\;
            Calculate $\mathcal{L}^{UM_v}$ and $\mathcal{L}^{UM_a}$ using the training samples $N$, respectively.\;
            Filter the noisy labels in each modality\;
            $N^{a}=\arg \min _{\tilde{N^{a}}:\left|\tilde{N^{a}}\right| \geq \tau\left|N\right|} \mathcal{L}^{UM_a}\left(N\right)$\;
            $N^{v}=\arg \min _{\tilde{N^{v}}:\left|\tilde{N^{v}}\right| \geq \tau\left|N\right|} \mathcal{L}^{UM_v}\left(N\right)$\\
            $N^{\prime}$=$N^v \bigcup N^a$\;
            Calculate $\mathcal{L}^{MM}$ using the training samples $N^{\prime}$\;
            Update $G^{MM}$ by $\mathcal{L}^{MM}$\;
            Update $G^{{UM}_a}$ by $\mathcal{L}^{UM_a}$\;  
            Update $G^{{UM}_v}$ by $\mathcal{L}^{UM_v}$\;
        }
    }
    \caption{Multi-modality Sample Cleaning}
    \label{alg:MMC}
\end{algorithm*}

\begin{figure*}[h]
    \hsize=\textwidth
    \centering
    \includegraphics[width=1\linewidth]{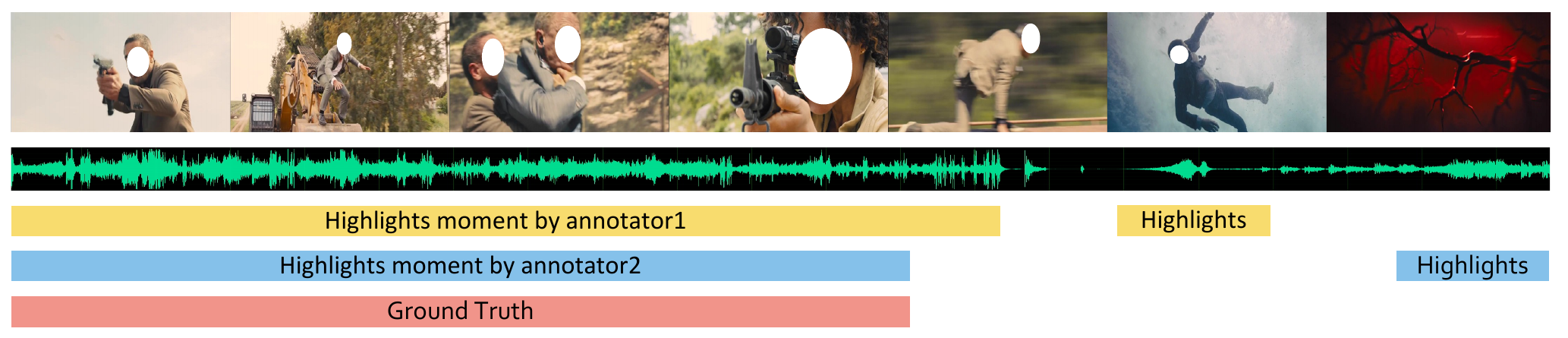}
    \captionof{figure}{
    Summary of MovieLights.}
    \vspace{10mm}
    \label{fig:summary}
\end{figure*}

\appendix
\section*{Appendix}

This supplemental material presents more details of our proposed CLC.
Appendix~ \ref{sec:mmc} describes the Multi-modality Sample Cleaning.
Appendix~\ref{sec:vis} lists some visualization examples and qualitative results in MoiveLights.
Finally, we present the broader impact in Appendix~\ref{sec:Broader_Impact}.


\section{Multi-modality Sample Cleaning}
\label{sec:mmc}

The detailed procedure of our MMC is presented in Algorithm~\ref{alg:MMC}.
The MMC module contains three branches, \emph{i.e.,} multi-modal branch $G^{MM}$, visual branch $G^{{UM}_v}$, and audio branch $G^{{UM}_a}$.
First, all instances in the two uni-modal branches are fed into the network to obtain the uni-modal losses, \emph{i.e.,} $\mathcal{L}^{UM_v}$ and $\mathcal{L}^{UM_a}$.
Next, we select a proportion of instances $N^{v}$ and $N^{a}$ that have small training losses in each branch independently. The number of instances is controlled by $\tau$.
Then, we combine the selected samples $N^{v}$ and $N^{a}$ and take them as clean samples to train the multi-modal branch. Assume that the multi-modal loss is denoted as $\mathcal{L}^{MM}$.
Finally, we aggregate all the losses in three branches to update the network parameters.
Through joint optimization, the multi-modal branch progressively attains more trustable labels that make learning more robust.

\begin{figure*}[h]
   \begin{center}
   \includegraphics[width=1\linewidth]{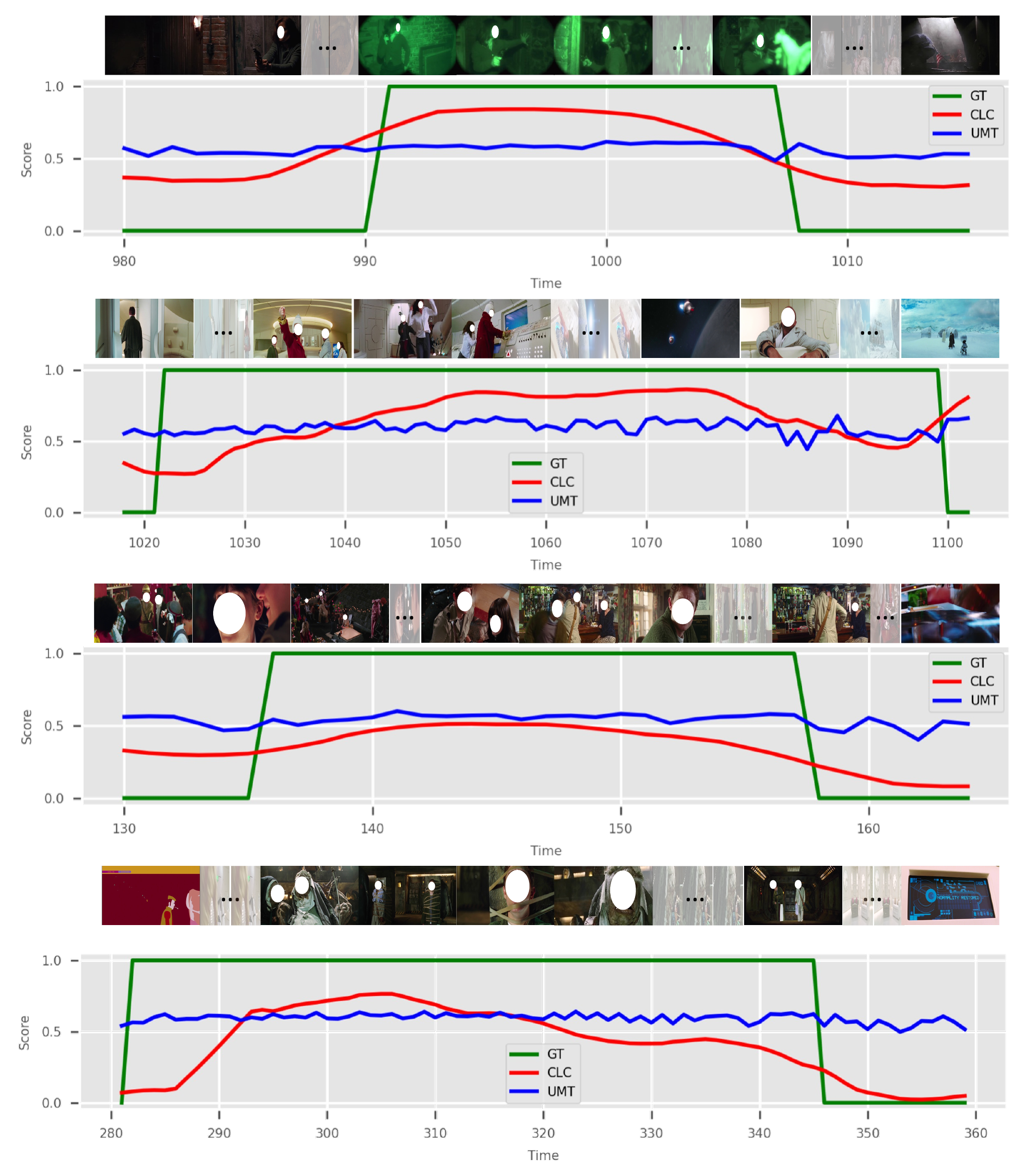}
   \end{center}
      \vspace{-0.5cm}
      \caption{Qualitative results. Prediction curve on MoivieLights detected by CLC and UMT. 
         }
      \vspace{-0.5cm}
   \label{fig:compare}
\end{figure*}

\begin{figure*}[h]
   \begin{center}
   \includegraphics[width=1\linewidth]{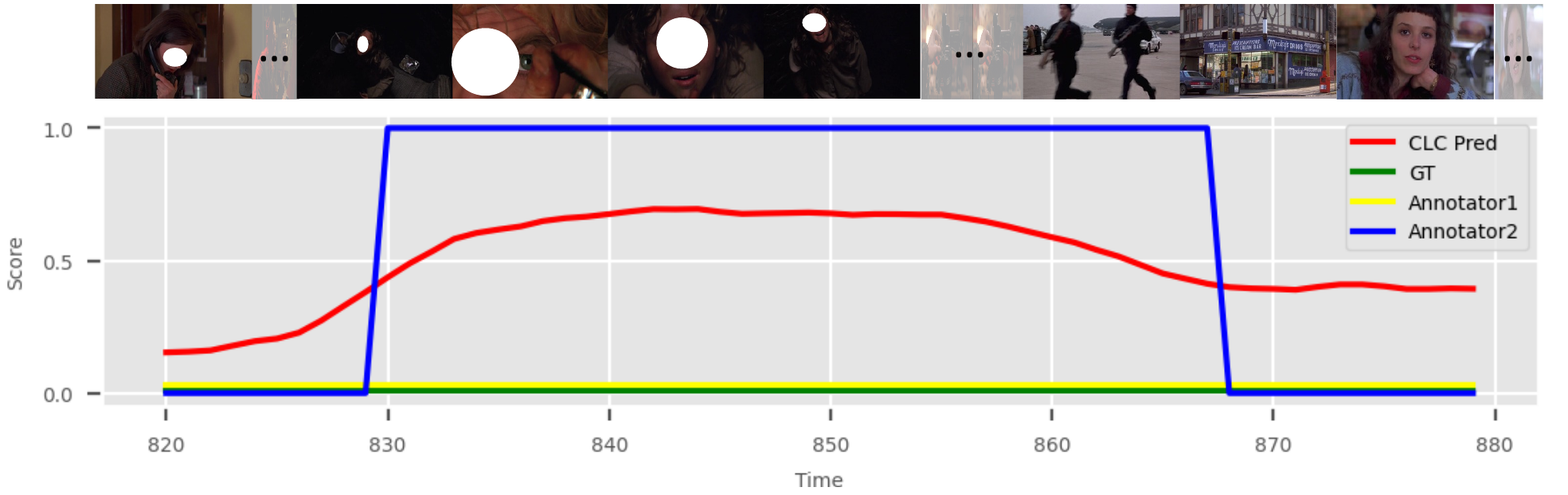}
   \end{center}
      \vspace{-0.5cm}
      \caption{
         Badcase on MoiveLights}
      \vspace{-0.5cm}
   \label{fig:badcase}
\end{figure*}

\section{Visualization}
\label{sec:vis}


As shown in Fig.~\ref{fig:summary}, we present a summary to introduce our MovieLights.
As seen in Fig.~\ref{fig:compare}, the predicted scores detected by CLC and UMT~\cite{UMTCVPR2022}  are shown by lines.
The results show that in different highlightness clips, the prediction score of UMT~\cite{UMTCVPR2022}  has a slight change, while the highlight moments and background scenes can be well distinguished by our CLC.
It indicates that our CLC is better at understanding movies and is more discriminative for highlights.
Fig.~\ref{fig:badcase} presents the badcase on MovieLights.
It incorrectly localized a highlight moment where one of the annotators regards it as a highlight and the other doesn't.
Since the subjectivity of the highlight detection, some highlights moment we detect maybe attract only a part of the audience.
It reflects the challenge of video highlight detection, and we hope to find better solutions in future work.



\section{Broader Impact} 
\label{sec:Broader_Impact}
With the growing number of new publications of movies and the rapid rise of short videos, it is necessary to train automatic movie highlight detection algorithms.
This work provides a scene-aware paradigm to learn highlight moments in movies without any manual annotation.
Besides, our  introduce a framework named Collaborative noisy Label Cleaner (CLC) to learn from these pseudo noisy labels. 
Finally, the collected dataset MovieLights could foster the further study of movie analysis.
The potential negative impact lies in that this dataset may be abused and may cause copyright issues. 
Hence, to avoid privacy and copyright issues, trailers and movies will be released in the form of extracted features in visual and audio modalities. 
If actual business data needs to be applied, it should be regulated and consented to by media providers.

{\small
\bibliographystyle{ieee_fullname}
\bibliography{egbib}
}

\end{document}